\pdfoutput=1
\PassOptionsToPackage{hyperfootnotes=false}{hyperref}
\documentclass[a4paper,fleqn]{cas-sc}

\usepackage[numbers,sort&compress]{natbib}
\usepackage{amsmath,amssymb}
\usepackage{booktabs}
\usepackage{graphicx}
\usepackage{hyperref}
\usepackage{xurl}

\ExplSyntaxOn
\cs_set:Npn \__first_footerline:
{ \group_begin: \small \sffamily \__short_authors: :~ { \rmfamily \itshape Preprint } \group_end: }
\ExplSyntaxOff

\begin{document}

\shorttitle{Cross-Layer VLA Optimisation Transfer}
\shortauthors{Huang et al.}

\title[mode=title]{Do Inference Optimisations Survive the Closed Loop?
A Cross-Layer Time-Budget Analysis of SmolVLA with Jetson Orin Nano Inference}

\author[1]{Yuhang Huang}

\author[1]{Yueheng Zhang}

\author[1]{Xinrui Shen}

\author[1]{Zhiqiang Que}
\cormark[1]

\cortext[cor1]{Corresponding author. Email address: z.que@bristol.ac.uk.}

\affiliation[1]{organization={School of Electrical, Electronic and Mechanical
Engineering, University of Bristol},
addressline={Queens Building, University Walk},
city={Bristol},
postcode={BS8 1TR},
country={United Kingdom}}

\begin{abstract}
Inference optimisations for vision--language--action (VLA) policies are often
measured at a component or policy-call boundary, although deployment outcomes
depend on whether those gains survive remote execution and the closed loop. We
trace optimisation transfer for one SmolVLA checkpoint across
fixed-shape components, policy calls, PC--Jetson requests and
LIBERO-Spatial episodes. The PC retains simulation and evaluation, while a
Jetson Orin Nano serves policy inference over HTTP; each closed-loop
configuration contains 50 episodes. Relative to the released
\((N,E,H)=(10,1,50)\) setting, the selected \((2,20,20)\) schedule reduced mean
PC-local episode time from 90.8 to 28.5~s (68.6\%), with observed success of
78\% versus 72\% and overlapping Wilson intervals. Weight-only post-training
quantisation reduced stored-parameter size by up to 35.3\% and peak allocation
by 25.0\%, but none of the configurations beat FP16 mean policy latency
in Jetson-hosted hardware-in-the-loop (HIL). A TensorRT connector recovered
3.1\% relative to language INT8, yet the integrated policy remained 3.5\%
slower than the FP16 full-policy baseline. Policy inference contributed
96.1--96.8\% of mean request latency, whereas completed remote requests
accounted for only 18.7--22.7\% of mean episode duration. Counterfactual trace
replay at \(E=20\) covered 395/395 FP16 requests but 0/423 mixed 4/8-bit
requests under a 1~s deadline. These results show that optimisation decisions
change with measurement boundary and request budget. They do not constitute
asynchronous control validation and are limited to one checkpoint, one edge
device and simulated tasks.
\end{abstract}


\maketitle

\section{Introduction}
Vision--language--action (VLA) policies map visual observations, language instructions and robot state to actions, making general-purpose robotic control possible with a shared policy \cite{rt2,openvla}. Smaller models such as SmolVLA lower the model-size barrier, yet iterative action generation and repeated closed-loop calls still impose latency and memory costs on constrained inference hardware \cite{smolvla}. The central deployment question is not simply whether one operator becomes faster, but whether that gain survives four progressively wider measurement boundaries: the accelerated component, the complete policy call, the remote request and the closed-loop episode.

Recent work accelerates VLA inference through fewer or adaptively selected refinement steps, asynchronous action execution, quantisation and model- or kernel-level optimisation \cite{snapflow,adavla,rtc,sqil,actquant,efficientvla}. These approaches use different models, training interventions, devices and timing boundaries. Consequently, their gains cannot by themselves establish whether an intervention remains beneficial when the same released policy moves from an isolated benchmark into a remote edge route and a simulated closed loop. Reducing policy-call cost may also change feedback frequency or task success, while reducing weight storage need not accelerate a runtime that dequantises weights before floating-point operations.

We test optimisation transfer for one pretrained SmolVLA checkpoint, without retraining. In the PC-local route, the PC performs policy inference and runs the LIBERO Spatial tasks. In the Jetson-hosted hardware-in-the-loop (HIL) route, the PC retains MuJoCo simulation, rendering and evaluation, while a Jetson Orin Nano serves policy inference over HTTP. Every reported closed-loop configuration covers ten tasks with five episodes per task. We compare fixed flow-refinement and action-execution schedules, component-wise weight-only post-training quantisation (PTQ), a TensorRT connector integration and a separate CUDA Graph component benchmark. We keep the component, policy-only, request and episode boundaries distinct, then replay retained request traces against explicit hypothetical deadlines.

We address three research questions. RQ1 asks which fixed inference-time schedule provides the most favourable observed trade-off between task success and episode time. RQ2 asks how weight-only PTQ affects parameter storage, peak allocation and policy latency on PC-local and Jetson-hosted routes. RQ3 asks which optimisation gains survive the transition from components to the complete policy, remote requests and the HIL closed loop. This paper contributes a boundary-aware evaluation rather than a new scheduling or quantisation algorithm. It identifies a practical fixed operating point, measures the memory--latency trade-off of a dequantise-then-compute PTQ path, and connects component reintegration with request-level accounting and counterfactual deadline coverage. Together, these results show where an apparent gain is retained, weakened or lost as the evaluation boundary widens.

\section{Related Work}
\subsection{Action-generation schedules and execution timing}
Generalist VLA policies span different action-generation designs, from RT-2 and OpenVLA to the compact SmolVLA policy examined here \cite{rt2,openvla,smolvla}. SmolVLA already analyses action-chunk length and the number of actions executed before a new observation, and introduces an asynchronous client--server stack that overlaps chunk generation with action execution \cite{smolvla}. RTC similarly addresses stalls at chunk boundaries through asynchronous execution \cite{rtc}. SnapFlow distils iterative flow matching into one-step generation, whereas AdaVLA changes refinement depth and MLP pruning online without retraining \cite{snapflow,adavla}. Adaptive Action Chunking and AutoHorizon instead adapt execution commitment from policy signals \cite{aac2026,autohorizon2026}. We therefore do not claim action chunking, execution-step selection or asynchronous control as contributions. The fixed \((N,E,H)\) sweep in this study is a controlled operating-point selection for the subsequent deployment analysis, not a new scheduling algorithm.

\subsection{Quantisation and edge VLA deployment}
SQIL combines quantisation-aware training with a saliency-weighted objective, whereas ActQuant applies action-guided mixed-precision PTQ with a native low-bit execution path \cite{sqil,actquant}. LiteVLA-Edge studies quantised on-device multimodal control on embedded hardware \cite{litevlaedge}. A broader XPU characterisation evaluates multiple VLA families across heterogeneous accelerators using cost, energy and time, and combines component profiling with control-rate screening and physical-robot validation \cite{xpu_characterization2026}. AdaVLA also reports LIBERO latency and energy on a Jetson AGX Orin, together with SmolVLA experiments on an SO-ARM101 robot \cite{adavla}. These studies establish a wider hardware and real-robot scope than the present work. Our narrower setting uses an entry-level Orin Nano as a remote policy server while the PC retains simulation, and tests a custom weight-only path that dequantises weights before floating-point linear operations. Storage and peak allocation are therefore reported separately from latency, and no native low-bit speed-up is inferred.

\subsection{Deployment characterisation across measurement boundaries}
EfficientVLA accelerates CogACT through architectural redundancy and feature reuse, while TensorRT engines and CUDA Graph replay target supported, often fixed-shape execution regions \cite{efficientvla,nvidia_tensorrt,nvidia_cuda_graphs}. A component-kernel gain alone does not establish a lower complete-policy latency or closed-loop benefit. This distinction is reinforced by a contemporaneous study of the same \nolinkurl{HuggingFaceVLA/smolvla_libero} checkpoint: ONNX Runtime reduced latency relative to its PyTorch+AMP route on an RTX 2060, but closed-loop Spatial success depended strongly on the exported interface, and the requested low-precision artifacts required graph-level verification \cite{onnx_smolvla2026}. That study covers two LIBERO suites and paired rollouts on a desktop GPU, but not a Jetson or remote PC--edge route.

The present study is consequently positioned as a bounded systems analysis rather than a new optimisation method or the broadest hardware characterisation. It follows one released checkpoint across PC-local scheduling, policy-only PTQ, PC--Jetson request latency, descriptive episode-time accounting and component reintegration. This evidence chain tests whether an apparent gain survives a change in measurement boundary. Its specific scope is the Orin Nano remote-inference route and its component-to-policy-to-episode transfer; it does not supersede broader XPU, energy, physical-robot or runtime-specific deployment studies.

\section{Cross-Layer Evaluation Methodology}
\subsection{System boundaries and closed-loop protocol}
\begin{figure}
  \centering
  \includegraphics[width=\textwidth]{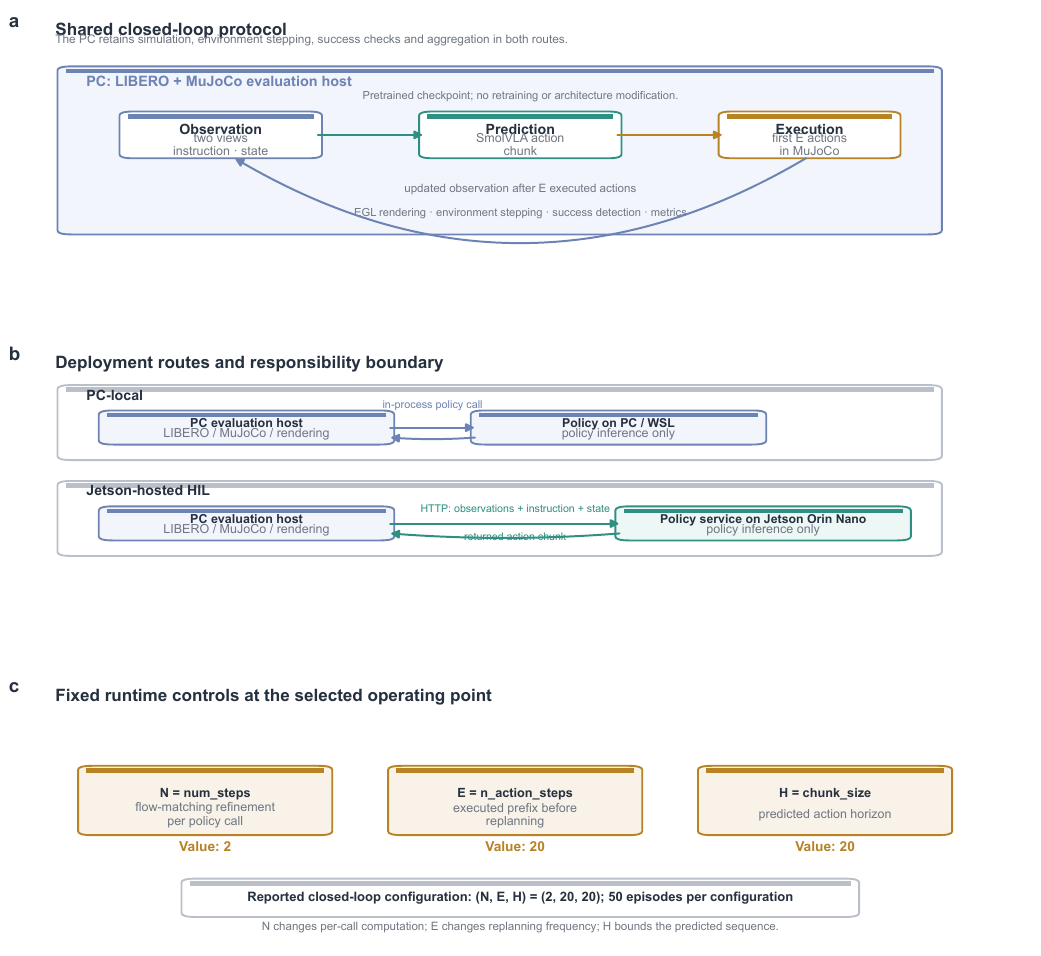}
  \caption{Controlled VLA routes and runtime protocol. \textbf{(a)} Shared observation--prediction--execution loop. \textbf{(b)} The PC retains simulation and evaluation in both routes; Jetson-hosted HIL moves policy inference only. \textbf{(c)} Runtime controls and the reported HIL point \((N,E,H)=(2,20,20)\), using the 50 episodes per closed-loop configuration.}
  \label{fig:system_protocol}
\end{figure}

All experiments use the pretrained \nolinkurl{HuggingFaceVLA/smolvla_libero} checkpoint at revision \nolinkurl{6721902bc4d61e50a3bfdb11dfb4cb626f05d102}, without retraining. Rollouts use WSL Ubuntu, Python 3.12.7, LeRobot 0.6.1, HF-LIBERO 0.1.4 and MuJoCo 3.3.2 with EGL rendering \cite{libero,mujoco}. The policy receives agent-view and wrist-view images, the task instruction and robot state. It returns a chunk of continuous actions, whose executed prefix advances the simulation before the next observation. Figure~\ref{fig:system_protocol} shows this common closed-loop protocol.

In the PC-local route, the PC runs policy inference through \texttt{lerobot-eval}, as well as LIBERO task execution, MuJoCo physics and rendering, environment stepping, success detection and metric aggregation. In the Jetson-hosted hardware-in-the-loop (HIL) route, the PC retains those simulator and evaluator functions and sends observations, instructions and robot state over HTTP to an NVIDIA Jetson Orin Nano (L4T R36.4.4) for policy inference; the returned action chunk is executed on the PC. The archived Jetson runtime records TensorRT 10.11.0, ONNX 1.17.0 and PyTorch \texttt{2.8.0a0+5228986}. The client serialises requests as HTTP/JSON, encoding the camera views as PNG images in base64 together with the instruction, 8-D state and previous action. CPU/GPU/RAM snapshots for the historical PC runs, Jetson memory SKU, power mode, clock state, link rate and a fixed action-rate setting were not archived; this study therefore does not report or infer them. The component-engine and PC policy-only benchmarks described below use separate measurement boundaries.

Each reported closed-loop configuration runs all ten LIBERO-Spatial tasks, with five episodes per task and an episode limit of 280 environment steps (50 episodes per configuration). The evaluator's binary success signal defines task success. We report mean episode time over the complete rollout set. Table~\ref{tab:protocol_boundaries} lists the routes and settings that determine which latency comparisons are valid.

\begin{table}
  \centering
  \caption{Protocol boundaries for the reported evidence.}
  \label{tab:protocol_boundaries}
  \small\setlength{\tabcolsep}{4pt}
  \renewcommand{\arraystretch}{1.18}
  \begin{tabular}{@{}p{0.24\textwidth}p{0.24\textwidth}p{0.41\textwidth}@{}}
    \toprule
    Route or measurement & Setting & Measurement boundary \\
    \midrule
    PC-local closed loop & 50 episodes; fixed schedules & PC runs policy, simulation and evaluation. \\
    Jetson-hosted HIL & \((2,20,20)\); 50 episodes & PC runs simulation; Jetson runs policy inference over HTTP. \\
    PC policy-only benchmark & \((2,1,50)\); 50 timed calls & CUDA-synchronised latency, storage and peak allocation. \\
    Component engines & Fixed input shape & CUDA-event latency and wall-clock throughput have distinct definitions. \\
    \bottomrule
  \end{tabular}
  \renewcommand{\arraystretch}{1}
\end{table}

\subsection{Fixed inference-time scheduling}
Let \(N=\texttt{num\_steps}\) be the number of flow-matching refinement iterations, \(H=\texttt{chunk\_size}\) the predicted action horizon and \(E=\texttt{n\_action\_steps}\) the executed prefix. For observation \(\mathbf{o}_t\), instruction \(\mathbf{l}\) and state \(\mathbf{s}_t\), the policy generates
\begin{equation}
\mathbf{A}_t=F_\theta(\mathbf{o}_t,\mathbf{l},\mathbf{s}_t;N,H)\in\mathbb{R}^{H\times d_a},
\label{eq:action_chunk}
\end{equation}
where \(d_a\) is the action dimension. The evaluator executes \(\mathbf{A}_t[1:E]\), with \(1\leq E\leq H\), before replanning. Thus \(N\) changes per-call refinement cost, \(E\) changes feedback frequency and \(H\) bounds the generated sequence.

The PC-local refinement sweep holds \((E,H)=(1,50)\) and varies \(N\in\{10,5,2\}\), using the released setting \((10,1,50)\) as its baseline. The execution-horizon sweep holds \((N,H)=(2,50)\) and varies \(E\in\{1,5,10,20,50\}\). A subsequent comparison holds \(N=2\) and matches \(H=E\) at \(E\in\{1,10,20\}\), against \(H=50\) at the same \(E\). These are fixed configurations: neither the task nor the state selects parameters during a rollout. The \((2,20,20)\) operating point was selected from the measured success--time trade-off and then held fixed for all HIL comparisons.

\subsection{Component-wise post-training quantisation}
Weight-only post-training quantisation (PTQ) replaces selected linear layers of the vision--language backbone after loading the pretrained checkpoint. Each weight matrix is partitioned along its input dimension into groups of 128 values, zero-padding the last group when needed. For bit width \(b\) and group \(\mathbf{w}_g\), symmetric absolute-maximum scaling is
\begin{equation}
s_g=\frac{\max\!\left(\max_i|w_{g,i}|,10^{-6}\right)}{2^{b-1}-1},\qquad
q_{g,i}=\operatorname{clip}\!\left(\operatorname{round}(w_{g,i}/s_g),-(2^{b-1}-1),2^{b-1}-1\right).
\label{eq:ptq_quantisation}
\end{equation}
INT8 weights are stored as signed integers with one FP32 scale per group; pairs of signed INT4 weights are packed into \texttt{uint8} storage. Quantised weights are dequantised to the input tensor type during each forward pass, followed by floating-point linear operations. Activation precision and the original mixed-precision execution path are retained. No calibration dataset, fine-tuning or gradient update is used.

The four reported configurations are FP16; language INT8 (text transformer only); backbone INT8 (vision encoder, connector and text transformer); and mixed 4/8-bit (INT4 vision encoder and connector, INT8 text transformer). The flow-matching action expert, embeddings and output heads remain FP16 in all cases. The timed PC policy-only benchmark fixes \((N,E,H)=(2,1,50)\). Figure~\ref{fig:ptq_tradeoff}a additionally reports archived PC-local closed-loop success records at released depth \(N=10\) for FP16, language INT8 and backbone INT8, and at candidate depth \(N=2\) for all four configurations. The archived source for those closed-loop rows does not independently record \(E\) and \(H\), so this paper does not use the \(N=10\) rows for an absolute latency comparison. Jetson HIL comparisons use the selected \((2,20,20)\) point. This custom dequantisation route does not execute the full policy with native low-bit arithmetic.

\subsection{Measurement, uncertainty and evidence boundaries}
Policy inference latency is the duration of a policy request at the inference host. In HIL, round-trip latency extends from the PC request to receipt of the action chunk; request-level transport/service overhead is \(T_{\mathrm{rt}}-T_{\mathrm{inf}}\), where \(T_{\mathrm{rt}}\) is round-trip latency and \(T_{\mathrm{inf}}\) is server-side policy-inference latency. Episode time is the wall-clock duration of a complete simulator rollout. We report mean policy latency and, where archived, p95 latency. For HIL traces, we also inspect p99 and maximum values from prediction-request records. Each HIL summary includes every logged prediction record because no exclusion criterion was specified. Success proportions use Wilson 95\% intervals over the 50 binary episode outcomes.

For each HIL configuration, we additionally relate the request trace to the aggregate episode duration through a descriptive time accounting. Let \(N_{\mathrm{req}}\) be the number of retained requests and \(N_{\mathrm{ep}}\) the number of episodes. The mean request contribution per episode is calculated directly from the total logged request time:
\begin{equation}
\bar T_{\mathrm{request}}=\frac{1}{N_{\mathrm{ep}}}\sum_{i=1}^{N_{\mathrm{req}}}T_{\mathrm{rt},i}
=\frac{N_{\mathrm{req}}}{N_{\mathrm{ep}}}\bar T_{\mathrm{rt}},\qquad
\bar T_{\mathrm{episode}}=\bar T_{\mathrm{request}}+T_{\mathrm{residual}},
\label{eq:episode_accounting}
\end{equation}
where \(\bar T_{\mathrm{episode}}\), \(N_{\mathrm{req}}\) and every \(T_{\mathrm{rt},i}\) are measured, and \(T_{\mathrm{residual}}\) is obtained by subtraction. This equality uses the request-time total, so it does not assume independence between per-episode request counts and latencies. The residual covers all time not represented by completed remote requests, including environment stepping, rendering, logging, request scheduling and other uninstrumented work; it is not an independently measured simulator-time term. The archived evaluation summaries identify \(\bar T_{\mathrm{episode}}\) as wall-clock episode duration but do not establish whether reset or initialisation is inside that timer, so this timer-boundary uncertainty is also left in the residual. Because only the aggregate mean episode duration was retained, Eq.~\eqref{eq:episode_accounting} is an accounting identity rather than a fitted or predictive timing model, and no episode-level interval is reported for its terms.

No asynchronous controller was evaluated, but the retained traces permit a counterfactual request-budget analysis. If a generated prefix of \(E\) actions were consumed at rate \(f\), it would provide \(E/f\) seconds before another prefix is needed. A necessary request-level no-stall condition and the corresponding trace coverage are
\begin{equation}
T_{\mathrm{rt},i}\leq \frac{E}{f},\qquad
C(f)=\frac{1}{N_{\mathrm{req}}}\sum_{i=1}^{N_{\mathrm{req}}}
\mathbb{1}\!\left[T_{\mathrm{rt},i}\leq \frac{E}{f}\right].
\label{eq:rate_budget}
\end{equation}
The configured control rate was not archived. We therefore evaluate \(C(f)\) for \(E=20\) at hypothetical rates of 5, 10 and 20~Hz, corresponding to 4, 2 and 1~s request deadlines; these scenarios are not measurements of the actual control rate. Every retained request, including the first, enters each coverage value. This trace replay does not model action-execution jitter, queue management, overlap or deadline consequences and is therefore neither a schedulability test nor evidence of asynchronous closed-loop operation. For descriptive tail auditing, we report the first request separately and call the distribution after removing that single record ``post-first-request'', not steady state.

The separate PC policy-only benchmark fixes \((N,E,H)=(2,1,50)\), executes five warm-up calls and measures 50 calls with \texttt{time.perf\_counter()} and CUDA synchronisation around each call. It records policy latency, stored-parameter size and peak allocated GPU memory. The FP16 TensorRT connector benchmark uses fixed 256- or 1,024-token inputs, 100 warm-up runs and 1,000 timed iterations; its CUDA-event latency excludes host--device transfer, while throughput uses synchronised wall-clock time. CUDA Graph replay is compared with eager enqueue on the same fixed-shape FP16 vision engine; enqueue time is recorded separately from GPU latency. These component measurements exclude the rest of the policy and the simulator loop. The incremental HIL ablation runs FP16, language INT8, and language INT8 plus an FP16 TensorRT connector under the same \((2,20,20)\), 50-episode protocol.

PC-local, Jetson HIL and component timings are compared within their respective routes and definitions. Most PC scheduling and PTQ results are available only as aggregate records. HIL episode outcomes and request traces are available for FP16 and the three PTQ configurations, but the archived metadata do not establish matching initial-state and inference-seed keys across configurations. The connector-integrated HIL condition has an aggregate summary but no per-episode or request-level trace. Accordingly, we report descriptive differences and Wilson intervals only: no episode-paired McNemar tests are applied, and no paired latency confidence intervals are given for the connector-integrated condition.

\section{Results}
\subsection{RQ1: fixed scheduling identifies a practical operating point}
\begin{figure}
  \centering
  \includegraphics[width=\textwidth]{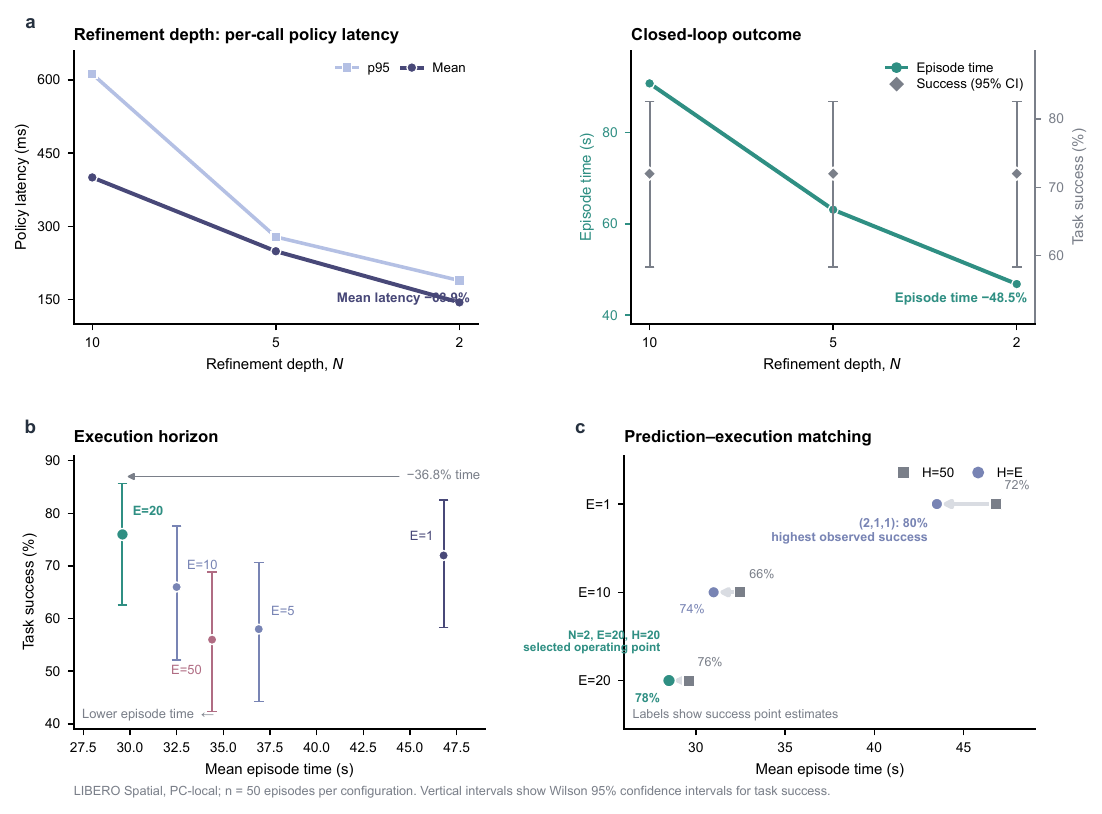}
  \caption{PC-local scheduling evidence. \textbf{(a)} Refinement-depth sweep. \textbf{(b)} Execution-horizon trade-off. \textbf{(c)} Matched prediction and execution horizons. Every point uses 50 LIBERO Spatial episodes; vertical intervals are Wilson 95\% intervals.}
  \label{fig:scheduling_tradeoff}
\end{figure}

A fixed reduction in refinement depth shortened PC-local episodes without changing the observed aggregate success point estimate in this sweep. Decreasing \(N\) from 10 to 2 reduced mean episode time from 90.8 to 46.8~s (Fig.~\ref{fig:scheduling_tradeoff}a; Table~\ref{tab:scheduling_outcomes}), while all three 50-episode configurations had 72\% observed success. Over the same range, mean policy-inference latency decreased from 400.4 to 144.6~ms.

The execution and prediction horizons exposed the feedback--runtime trade-off. At \(N=2\), increasing the executed prefix to \(E=20\) reduced mean episode time to 29.6~s, with 76\% observed success (Fig.~\ref{fig:scheduling_tradeoff}b). Executing the full horizon (\(E=50\)) instead lengthened mean episode time to 34.4~s, relative to \(E=20\), and lowered observed success to 56\%. Matching the prediction and execution horizons at \((N,E,H)=(2,20,20)\) produced the lowest observed mean episode time among the tested fixed settings, 28.5~s, with 78\% observed success (Fig.~\ref{fig:scheduling_tradeoff}c). Relative to the released \((10,1,50)\) setting, this selected point reduced mean episode time by 68.6\%. The two observed success estimates were 78\% and 72\%, with overlapping Wilson intervals. We use this fixed operating point in the subsequent HIL experiments.

\begin{table}
  \centering
  \caption{PC-local scheduling outcomes; each row uses 50 episodes. Policy latency is mean/p95 in ms where archived; -- denotes unavailable measurements.}
  \label{tab:scheduling_outcomes}
  \scriptsize\setlength{\tabcolsep}{3pt}
  \begin{tabular}{@{}llccc@{}}
    \toprule
    Experiment & \((N,E,H)\) & Success (\%) & Episode (s) & Policy mean/p95 (ms) \\
    \midrule
    Refinement & (10,1,50) & 72 & 90.8 & 400.4 / 612.1 \\
    Refinement & (5,1,50) & 72 & 63.1 & 249.3 / 278.6 \\
    Refinement & (2,1,50) & 72 & 46.8 & 144.6 / 189.4 \\
    Execution horizon & (2,20,50) & 76 & 29.6 & -- \\
    Execution horizon & (2,50,50) & 56 & 34.4 & -- \\
    Matched horizon & (2,20,20) & 78 & 28.5 & -- \\
    \bottomrule
  \end{tabular}
\end{table}

\subsection{RQ2: PTQ transfers memory savings, but not Jetson HIL latency gains}
\begin{figure}
  \centering
  \includegraphics[width=\textwidth]{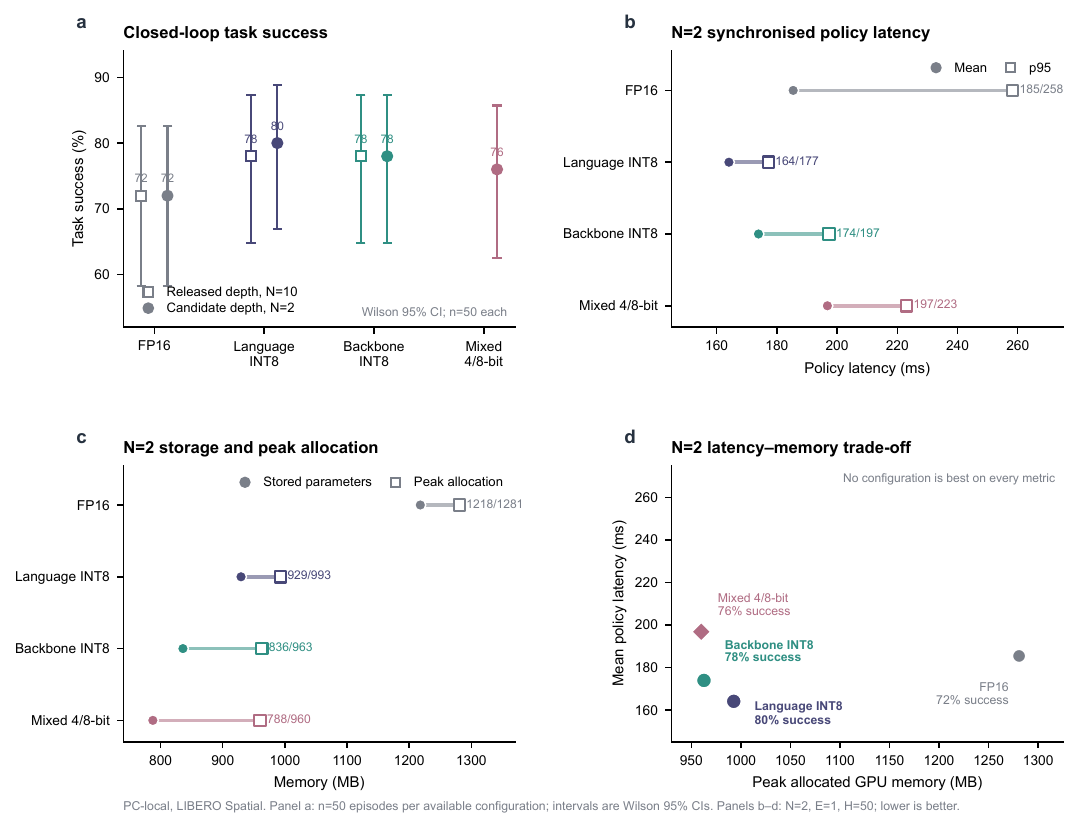}
  \caption{Cross-layer PTQ evidence. \textbf{(a)} PC-local closed-loop success from 50 episodes per available configuration; released-depth \(N=10\) points are unavailable for mixed 4/8-bit. \textbf{(b,c)} Separate CUDA-synchronised policy-only latency and memory at \((N,E,H)=(2,1,50)\). \textbf{(d)} Latency--memory trade-off; no configuration is best on every metric.}
  \label{fig:ptq_tradeoff}
\end{figure}

Weight-only PTQ consistently reduced parameter storage and peak memory, but did not produce a route-independent latency gain. In the separate PC policy-only benchmark, language INT8 reduced stored parameters from 1217.9 to 929.4~MB and peak allocation from 1280.8 to 992.9~MB, while yielding the lowest measured mean/p95 latency of 164.1/177.2~ms (Fig.~\ref{fig:ptq_tradeoff}b--d; Table~\ref{tab:ptq_cross_layer}a). Backbone INT8 and mixed 4/8-bit further reduced stored parameters to 835.9 and 787.5~MB, respectively. Mixed 4/8-bit also had the lowest peak allocation, 960.1~MB, but the highest mean latency, 196.8~ms, although its p95 latency (223.1~ms) remained below that of FP16 (258.3~ms). In PC-local closed-loop evaluation at \(N=2\), the four configurations had observed success rates from 72\% to 80\%, with overlapping Wilson intervals (Fig.~\ref{fig:ptq_tradeoff}a).

The HIL route did not preserve the PC policy-only latency ordering. At the common \((2,20,20)\) setting, the Jetson-hosted FP16 configuration had the lowest mean policy-inference latency, 825.1~ms. Language INT8, backbone INT8 and mixed 4/8-bit measured 881.8, 938.1 and 1012.5~ms, respectively (Table~\ref{tab:ptq_cross_layer}b). The corresponding success point estimates ranged from 66\% to 82\% with 50 episodes per configuration, and their Wilson intervals overlapped. These observations support a memory--latency trade-off within the tested custom dequantisation path. They do not identify hardware-native low-bit acceleration or permit direct comparisons between PC policy-only and Jetson HIL latencies.

\begin{table}
  \centering
  \caption{Cross-layer PTQ evidence. PC policy-only and Jetson HIL are different routes.}
  \label{tab:ptq_cross_layer}
  \scriptsize\setlength{\tabcolsep}{2.5pt}\renewcommand{\arraystretch}{1.12}
  \noindent\textbf{a \quad PC policy-only: \((N,E,H)=(2,1,50)\)}\par\smallskip
  \begin{tabular*}{\linewidth}{@{\extracolsep{\fill}}lrrrr@{}}
    \toprule Configuration & Mean (ms) & p95 (ms) & Stored (MB) & Peak (MB) \\ \midrule
    FP16 & 185.4 & 258.3 & 1217.9 & 1280.8 \\
    Language INT8 & 164.1 & 177.2 & 929.4 & 992.9 \\
    Backbone INT8 & 173.9 & 197.4 & 835.9 & 962.8 \\
    Mixed 4/8-bit & 196.8 & 223.1 & 787.5 & 960.1 \\ \bottomrule
  \end{tabular*}\par\medskip
  \noindent\textbf{b \quad Jetson HIL: \((N,E,H)=(2,20,20)\), 50 episodes}\par\smallskip
  \begin{tabular*}{\linewidth}{@{\extracolsep{\fill}}lrrr@{}}
    \toprule Configuration & Success (\%) & Inference mean (ms) & Episode time (s) \\ \midrule
    FP16 & 72 & 825.1 & 36.2 \\
    Language INT8 & 80 & 881.8 & 33.3 \\
    Backbone INT8 & 82 & 938.1 & 33.5 \\
    Mixed 4/8-bit & 66 & 1012.5 & 38.9 \\ \bottomrule
  \end{tabular*}\par
\end{table}

\subsection{RQ3: optimisation gains weaken as the measurement boundary widens}
\begin{figure}
  \centering
  \includegraphics[width=\textwidth]{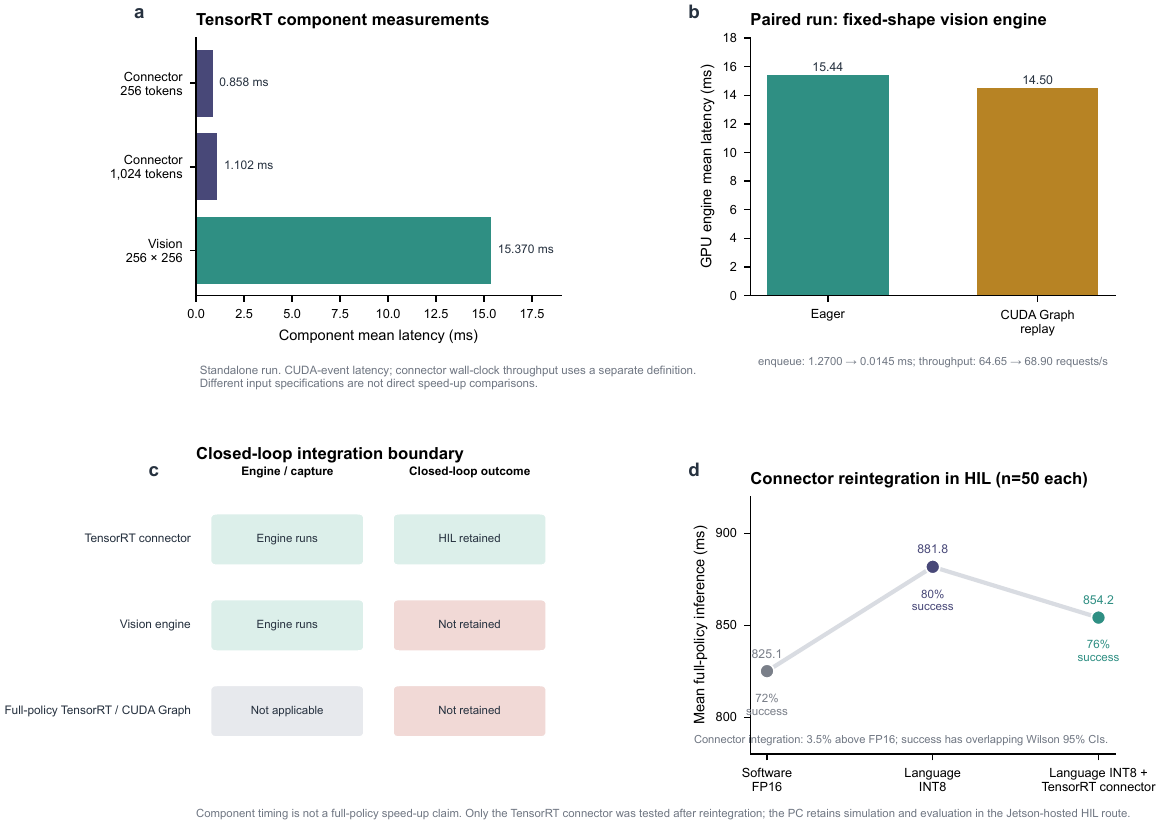}
  \caption{Component-to-closed-loop evidence. \textbf{(a)} TensorRT component timing with distinct protocols. \textbf{(b)} Paired eager and CUDA Graph replay timing of the fixed-shape FP16 vision engine. \textbf{(c)} Integration boundary. \textbf{(d)} HIL connector ablation using 50 episodes per configuration. Component gains are not interpreted as full-policy speed-ups.}
  \label{fig:component_to_closed_loop}
\end{figure}

The component benchmarks established executable engine paths, but not a faster complete policy. The FP16 TensorRT connector engines measured 0.858~ms for a 256-token input and 1.102~ms for a 1,024-token input using CUDA Events. A separate 256~\(\times\)~256 FP16 vision-engine probe measured 15.37~ms. In a distinct paired CUDA Graph run, the eager vision-engine baseline was 15.44~ms and replay measured 14.50~ms (Fig.~\ref{fig:component_to_closed_loop}a,b; Table~\ref{tab:jetson_integration_audit}). No matched eager-versus-TensorRT connector baseline is available in the archived evidence. Connector CUDA-event latency and synchronised wall-clock throughput have different definitions, and the connector rows use different token lengths. These results therefore establish component execution, not matched full-policy speed-up.

Full-policy reintegration narrowed the gain further. Adding the 1,024-token TensorRT connector to language INT8 reduced mean HIL policy-inference latency from 881.8 to 854.2~ms (3.1\%), but the integrated policy remained 3.5\% slower than the 825.1~ms FP16 HIL baseline (Fig.~\ref{fig:component_to_closed_loop}d). The three 50-episode success point estimates, 72\%, 80\% and 76\%, had overlapping Wilson intervals. The evidence supports connector compatibility and partial latency recovery, not a fully compiled SmolVLA runtime or a faster complete policy than FP16.

\begin{figure}
  \centering
  \includegraphics[width=\textwidth]{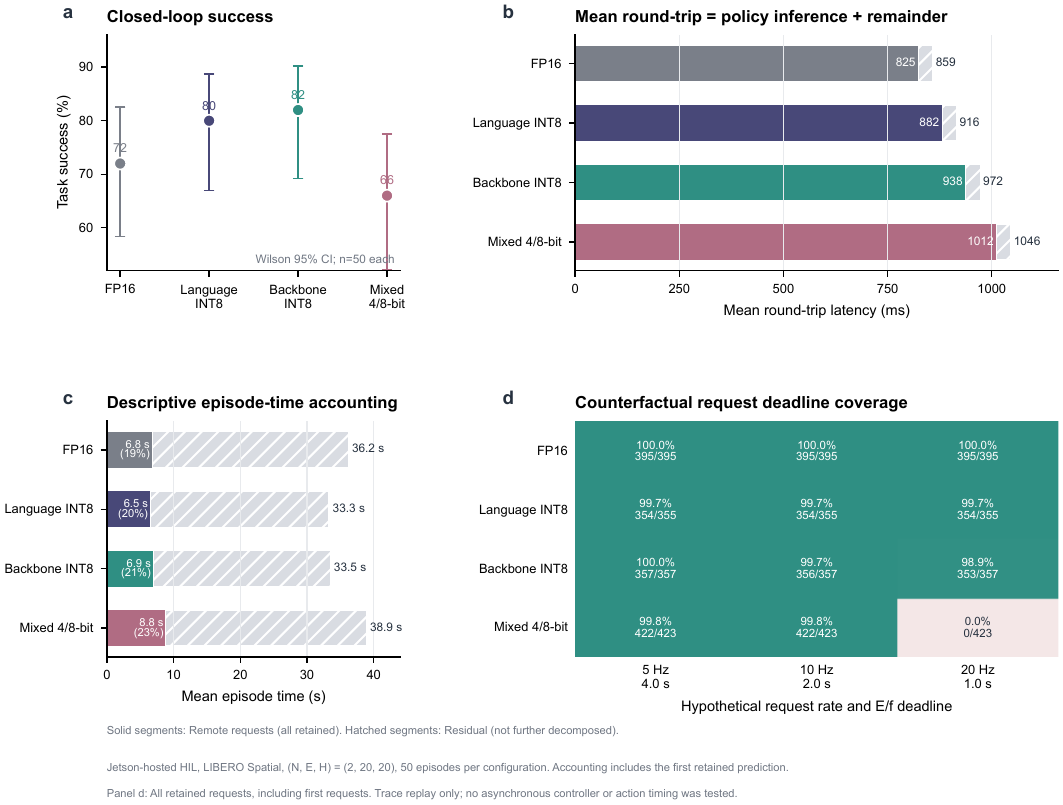}
  \caption{Jetson-hosted HIL evidence and cross-layer timing account. \textbf{(a)} Success with Wilson 95\% intervals. \textbf{(b)} Measured mean round-trip latency decomposed into policy inference and the remaining request-level time. \textbf{(c)} Descriptive mean episode-time accounting from Eq.~\eqref{eq:episode_accounting}; the hatched residual is obtained by subtraction and is not further decomposed. \textbf{(d)} Fraction and count of all retained requests satisfying the counterfactual \(E/f\) deadline for \(E=20\) at 5, 10 and 20~Hz. Panel~(d) is trace replay, not an asynchronous schedulability or closed-loop test. All panels use 50 episodes and retain the first prediction request.}
  \label{fig:jetson_hil_evidence}
\end{figure}

At the remote-request boundary, server-side policy inference dominated the measured round trip. Across all four configurations, inference accounted for 96.1--96.8\% of mean round-trip latency, while trace-derived transport and service overhead was 33.5--34.1~ms per request (Fig.~\ref{fig:jetson_hil_evidence}b). This attribution is calculated per request rather than from differences between summary quantiles. Each quantised trace contains one substantially longer first retained prediction request, labelled episode~0, step~20. The client sends that request with step 0 and records step 20 only after incrementing by the 20-action prefix. Because no exclusion rule was specified, these observations remain in the all-request means, p99 values and maxima reported in Table~\ref{tab:jetson_integration_audit}.

The aggregate accounting in Eq.~\eqref{eq:episode_accounting} assigns 6.5--8.8~s per episode to completed remote requests, or 18.7--22.7\% of the measured 33.3--38.9~s mean episode durations (Fig.~\ref{fig:jetson_hil_evidence}c). The calculated residual is 26.6--30.1~s per episode. Thus, policy inference is the largest measured component within a request, but remote requests explain only about one fifth of complete episode time under this synchronous HIL protocol. The residual cannot be divided further with the retained instrumentation and is not labelled as simulator time.

Counterfactual request budgets separated configurations that appeared similar under looser deadlines. In the trace replay of Eq.~\eqref{eq:rate_budget}, FP16 met all three deadlines for 395/395 retained requests (Fig.~\ref{fig:jetson_hil_evidence}d). Language INT8 met 354/355 deadlines at each hypothetical rate. Backbone INT8 met 357/357, 356/357 and 353/357 deadlines at 5, 10 and 20~Hz, respectively. Mixed 4/8-bit met 422/423 at 5 and 10~Hz but 0/423 at 20~Hz because every retained round trip exceeded the 1~s budget. The first retained requests were 4.31, 3.49 and 4.90~s for language INT8, backbone INT8 and mixed 4/8-bit; the corresponding post-first-request p99 values were 0.938, 0.984 and 1.064~s (FP16: 0.889~s). The actual control frequency is unknown, and no asynchronous deadline or closed-loop behaviour was tested.

\begin{table}
  \centering
  \caption{Jetson integration audit. HIL trace rows use \((2,20,20)\), 50 episodes, and include every logged prediction record.}
  \label{tab:jetson_integration_audit}
  \scriptsize\setlength{\tabcolsep}{2.0pt}\renewcommand{\arraystretch}{1.12}
  \noindent\textbf{a \quad Request-level HIL tail audit}\par\smallskip
  \begin{tabular*}{\linewidth}{@{\extracolsep{\fill}}lrrrp{0.26\linewidth}@{}}
    \toprule Configuration & Requests & Policy p99 / max (ms) & Overhead mean (ms) & \shortstack[l]{Location of\\maximum} \\ \midrule
    FP16 & 395 & 854.4 / 859.8 & 33.5 & -- \\
    Language INT8 & 355 & 905.3 / 4232.1 & 33.9 & episode 0, first logged record (step 20) \\
    Backbone INT8 & 357 & 947.0 / 3425.4 & 34.1 & episode 0, first logged record (step 20) \\
    Mixed 4/8-bit & 423 & 1032.5 / 4787.9 & 33.5 & episode 0, first logged record (step 20) \\ \bottomrule
  \end{tabular*}\par\medskip
  \noindent\textbf{b \quad Component protocol and retained closed-loop ablation}\par\smallskip
  \begin{tabular}{@{}p{0.27\linewidth}p{0.32\linewidth}rp{0.20\linewidth}@{}}
    \toprule Item & Measurement boundary & Mean latency (ms) & Closed-loop status \\ \midrule
    Connector probe & FP16; 256 tokens; CUDA Events & 0.858 & Component only \\
    Connector, 1,024-token input & FP16; CUDA Events & 1.102 & Reintegrated \\
    Vision-engine probe & FP16; 256 \(\times\) 256; standalone run & 15.37 & Component only \\
    CUDA Graph pair & FP16 vision engine; paired eager \(\rightarrow\) replay & 15.44 \(\rightarrow\) 14.50 & Component only \\
    HIL FP16 / Language INT8 / Language INT8 + TensorRT connector & Full policy; 50 episodes & 825.1 / 881.8 / 854.2 & 72 / 80 / 76\% success \\ \bottomrule
  \end{tabular}\par
\end{table}

\section{Discussion and Limitations}
\subsection{Measurement boundary changes the optimisation decision}
The main systems implication is that an optimisation should be selected at the widest boundary it is intended to improve. Refinement depth directly changed per-call work, but the executed prefix also changed feedback frequency and the number of policy requests. The resulting \((2,20,20)\) setting is therefore a practical operating point for this protocol, not a universally optimal schedule. Likewise, weight-only PTQ preserved storage and peak-allocation savings across measurement routes, while its latency ranking changed between the PC policy-only benchmark and Jetson HIL. A component or policy-only result is thus useful for screening candidates, but insufficient for selecting a closed-loop configuration. This boundary-aware interpretation complements training-based reductions in flow-matching steps \cite{snapflow} rather than treating fixed inference settings as a new scheduling method.

\subsection{Request budgets expose a different constraint from episode time}
Request latency and episode time answer different deployment questions. Policy inference formed almost all measured request latency, yet completed requests accounted for only about one fifth of synchronous HIL episode duration. The remaining time is not a measured mechanism; it combines all uninstrumented work and may also absorb uncertainty in the reset and initialisation timer boundary. Consequently, faster policy requests can be important for avoiding action-buffer depletion without producing a proportional reduction in simulated episode time. The counterfactual replay makes this distinction concrete: mixed 4/8-bit covered nearly every retained request at a 2~s deadline but none at 1~s. SmolVLA and RTC use asynchronous execution to address stalls at action-chunk boundaries \cite{smolvla,rtc}; the present analysis supplies a request-level screening criterion for such a design, not evidence that asynchronous deadline compliance or closed-loop behaviour has been validated.

\subsection{Integration constraints determine whether component gains transfer}
Component engines should be treated as compatibility evidence until the complete policy is reintegrated. CUDA Graph replay reduced latency for one fixed-shape vision engine, and the TensorRT connector partially recovered the language-INT8 policy latency, but neither result established a faster FP16-relative full policy. Fixed shapes, unsupported operators, tensor conversions and runtime dispatch may all restrict transfer, while the tested PTQ path adds forward-time dequantisation before floating-point linear operations. The retained measurements do not isolate these mechanisms, so the Jetson ordering should not be generalised to quantisation-aware training or native low-bit kernels \cite{sqil,actquant}. The reusable procedure is instead to benchmark a component, reintegrate it, measure the complete policy and then assess its request and closed-loop consequences.

\subsection{Threats to validity}
The empirical scope is one pretrained checkpoint, ten LIBERO Spatial simulation tasks, five episodes per task and one Jetson device. The PC still performs rendering, physics, environment control and evaluation, so the HIL route does not measure a fully on-device robot. No power, thermal, sustained-load or physical-robot transfer measurements are available. The scheduling operating point and PTQ ordering may change with another policy, task distribution, precision runtime or device.

The archived records also limit statistical comparison between configurations. PC scheduling and PTQ outcomes are largely aggregate, while HIL episode and request traces do not establish cross-configuration matching of initial states and inference seeds. Equal or overlapping success estimates therefore do not establish equivalence, and paired episode tests are not justified. The connector-integrated HIL condition lacks per-episode and request-level traces, so its latency recovery remains an aggregate comparison. The first retained request is included in the main latency summaries; removing it for a post-first-request audit does not establish stationarity. The episode summaries do not preserve the reset and initialisation timer boundary, and the actual action rate was not archived. These constraints prevent causal decomposition of the residual and validation of asynchronous deadline compliance.

\section{Conclusion}
We traced fixed scheduling, weight-only PTQ and component engines across component, policy, remote-request and simulated closed-loop boundaries for one pretrained SmolVLA checkpoint. The selected \((2,20,20)\) schedule reduced mean PC-local episode time by 68.6\% relative to the released setting, whereas PTQ transferred memory savings but not an FP16-relative Jetson HIL latency gain. TensorRT connector reintegration partially recovered language-INT8 latency but remained slower than the FP16 full-policy baseline. Request-level accounting further showed why the evaluation boundary matters: policy inference dominated each remote request, while completed requests formed only a minority of synchronous episode duration and deadline coverage changed sharply with the assumed budget. The supported conclusion is therefore methodological and bounded: VLA optimisations should be judged after reintegration at the measurement boundary they are intended to improve. The evidence does not establish asynchronous control performance and is limited to simulated tasks, one checkpoint and one Jetson device.

\bibliographystyle{cas-model2-unsrt}
\bibliography{references}

@misc{rt2,
  author = {Anthony Brohan and Noah Brown and Justice Carbajal and Yevgen Chebotar and Xi Chen and Krzysztof Choromanski and Tianli Ding and Danny Driess and Avinava Dubey and Chelsea Finn and Pete Florence and Chuyuan Fu and Montse Gonzalez Arenas and Keerthana Gopalakrishnan and Kehang Han and Karol Hausman and Alexander Herzog and Jasmine Hsu and Brian Ichter and Alex Irpan and Nikhil Joshi and Ryan Julian and Dmitry Kalashnikov and Yuheng Kuang and Isabel Leal and Lisa Lee and Tsang-Wei Edward Lee and Sergey Levine and Yao Lu and Henryk Michalewski and Igor Mordatch and Karl Pertsch and Kanishka Rao and Krista Reymann and Michael Ryoo and Grecia Salazar and Pannag Sanketi and Pierre Sermanet and Jaspiar Singh and Anikait Singh and Radu Soricut and Huong Tran and Vincent Vanhoucke and Quan Vuong and Ayzaan Wahid and Stefan Welker and Paul Wohlhart and Jialin Wu and Fei Xia and Ted Xiao and Peng Xu and Sichun Xu and Tianhe Yu and Brianna Zitkovich},
  title = {{RT-2}: Vision-Language-Action Models Transfer Web Knowledge to Robotic Control},
  year = {2023},
  eprint = {2307.15818},
  archivePrefix = {arXiv},
  primaryClass = {cs.RO},
  doi = {10.48550/arXiv.2307.15818},
  url = {https://arxiv.org/abs/2307.15818}
}

@misc{openvla,
  author = {Moo Jin Kim and Karl Pertsch and Siddharth Karamcheti and Ted Xiao and Ashwin Balakrishna and Suraj Nair and Rafael Rafailov and Ethan Foster and Grace Lam and Pannag Sanketi and Quan Vuong and Thomas Kollar and Benjamin Burchfiel and Russ Tedrake and Dorsa Sadigh and Sergey Levine and Percy Liang and Chelsea Finn},
  title = {{OpenVLA}: An Open-Source Vision-Language-Action Model},
  year = {2024},
  eprint = {2406.09246},
  archivePrefix = {arXiv},
  primaryClass = {cs.RO},
  doi = {10.48550/arXiv.2406.09246},
  url = {https://arxiv.org/abs/2406.09246}
}

@misc{smolvla,
  author = {Shukor, Mustafa and Aubakirova, Dana and Capuano, Francesco and Kooijmans, Pepijn and Palma, Steven and Zouitine, Adil and Aractingi, Michel and Pascal, Caroline and Russi, Martino and Marafioti, Andres and Alibert, Simon and Cord, Matthieu and Wolf, Thomas and Cadene, Remi},
  title = {{SmolVLA}: A Vision-Language-Action Model for Affordable and Efficient Robotics},
  year = {2025},
  eprint = {2506.01844},
  archivePrefix = {arXiv},
  primaryClass = {cs.LG},
  doi = {10.48550/arXiv.2506.01844},
  url = {https://arxiv.org/abs/2506.01844}
}

@misc{xpu_characterization2026,
  author = {Kaijun Zhou and Qiwei Chen and Da Peng and Zhiyang Li and Xijun Li and Jinyu Gu},
  title = {Characterizing Vision-Language-Action Models across {XPUs}: Constraints and Acceleration for On-Robot Deployment},
  year = {2026},
  eprint = {2604.24447},
  archivePrefix = {arXiv},
  primaryClass = {cs.RO},
  doi = {10.48550/arXiv.2604.24447},
  url = {https://arxiv.org/abs/2604.24447}
}

@misc{onnx_smolvla2026,
  author = {Rafiqul Islam},
  title = {When Faster {VLA} Deployment Changes Closed-Loop Behavior: Task Success--Latency Analysis of {SmolVLA} Across {PyTorch} and {ONNX} Variants},
  year = {2026},
  eprint = {2609.14146},
  archivePrefix = {arXiv},
  primaryClass = {cs.RO},
  doi = {10.48550/arXiv.2609.14146},
  url = {https://arxiv.org/abs/2609.14146}
}

@misc{snapflow,
  author = {Wuyang Luan and Junhui Li and Weiguang Zhao and Wenjian Zhang and Tieru Wu and Rui Ma},
  title = {{SnapFlow}: One-Step Action Generation for Flow-Matching {VLAs} via Progressive Self-Distillation},
  year = {2026},
  eprint = {2604.05656},
  archivePrefix = {arXiv},
  primaryClass = {cs.CV},
  doi = {10.48550/arXiv.2604.05656},
  url = {https://arxiv.org/abs/2604.05656}
}

@misc{adavla,
  author = {Sunghwan Han and Youngtae Han and Youngmin Yi},
  title = {{AdaVLA}: Adaptive Step Flow Matching for Training-free Acceleration of Vision-Language-Action Models},
  year = {2026},
  eprint = {2608.29208},
  archivePrefix = {arXiv},
  primaryClass = {cs.RO},
  doi = {10.48550/arXiv.2608.29208},
  url = {https://arxiv.org/abs/2608.29208}
}

@misc{sqil,
  author = {Seongmin Park and Hyungmin Kim and Sangwoo Kim and Wonseok Jeon and Juyoung Yang and Byeongwook Jeon and Yoonseon Oh and Jungwook Choi},
  title = {Saliency-Aware Quantized Imitation Learning for Efficient Robotic Control},
  year = {2025},
  eprint = {2505.15304},
  archivePrefix = {arXiv},
  primaryClass = {cs.RO},
  doi = {10.48550/arXiv.2505.15304},
  url = {https://arxiv.org/abs/2505.15304}
}

@misc{actquant,
  author = {Arash Akbari and Arman Akbari and Masih Eskandar and Qitao Tan and Yixiao Chen and Jingwu Luo and Bertha Pangaribuan and Liyun Zhang and Jennifer Dy and Geng Yuan and Xue Lin and Gaowen Liu and Stratis Ioannidis and Yanzhi Wang},
  title = {{ActQuant}: Sub-4-bit Action-Guided Quantization for Vision-Language-Action Models},
  year = {2026},
  eprint = {2605.24011},
  archivePrefix = {arXiv},
  primaryClass = {cs.CV},
  doi = {10.48550/arXiv.2605.24011},
  url = {https://arxiv.org/abs/2605.24011}
}

@misc{nvidia_tensorrt,
  author = {{NVIDIA Corporation}},
  title = {{NVIDIA TensorRT Documentation}},
  year = {2026},
  url = {https://docs.nvidia.com/deeplearning/tensorrt/latest/index.html},
  note = {Accessed: 2026-08-22}
}

@misc{nvidia_cuda_graphs,
  author = {{NVIDIA Corporation}},
  title = {{CUDA Programming Guide}: {CUDA Graphs}},
  year = {2026},
  url = {https://docs.nvidia.com/cuda/cuda-programming-guide/04-special-topics/cuda-graphs.html},
  note = {Accessed: 2026-08-22}
}

@inproceedings{efficientvla,
  author = {Yang, Yantai and Wang, Yuhao and Wen, Zichen and Luo, Zhongwei and Zou, Chang and Zhang, Zhipeng and Wen, Chuan and Zhang, Linfeng},
  title = {{EfficientVLA}: Training-Free Acceleration and Compression for Vision-Language-Action Models},
  year = {2025},
  booktitle = {Advances in Neural Information Processing Systems 38 (NeurIPS 2025)},
  volume = {38},
  pages = {45758--45781},
  doi = {10.52202/085713-1365},
  url = {https://proceedings.neurips.cc/paper_files/paper/2025/hash/3a2ef31a1e45908901adc0ca853a8faf-Abstract-Conference.html}
}

@inproceedings{rtc,
  author = {Black, Kevin and Galliker, Manuel and Levine, Sergey},
  title = {Real-Time Execution of Action Chunking Flow Policies},
  year = {2025},
  booktitle = {Advances in Neural Information Processing Systems 38 (NeurIPS 2025)},
  volume = {38},
  pages = {33383--33407},
  doi = {10.52202/085713-1122},
  url = {https://proceedings.neurips.cc/paper_files/paper/2025/hash/300ccb2187dedd4edcc07f7e76d8e553-Abstract-Conference.html}
}

@misc{aac2026,
  author = {Yuanchang Liang and Xiaobo Wang and Kai Wang and Shuo Wang and Xiaojiang Peng and Haoyu Chen and David Kim Huat Chua and Prahlad Vadakkepat},
  title = {Adaptive Action Chunking at Inference-time for Vision-Language-Action Models},
  year = {2026},
  eprint = {2604.04161},
  archivePrefix = {arXiv},
  primaryClass = {cs.RO},
  doi = {10.48550/arXiv.2604.04161},
  url = {https://arxiv.org/abs/2604.04161}
}

@misc{autohorizon2026,
  author = {Haoxuan Wang and Gengyu Zhang and Yan Yan and Ramana Rao Kompella and Gaowen Liu},
  title = {{VLA} Knows Its Limits: Adaptive Execution Horizons for Robot Policies},
  year = {2026},
  eprint = {2602.21445},
  archivePrefix = {arXiv},
  primaryClass = {cs.RO},
  doi = {10.48550/arXiv.2602.21445},
  url = {https://arxiv.org/abs/2602.21445}
}

@misc{litevlaedge,
  author = {Justin Williams and Kishor Datta Gupta and Roy George and Mrinmoy Sarkar},
  title = {{LiteVLA-Edge}: Quantized On-Device Multimodal Control for Embedded Robotics},
  year = {2026},
  eprint = {2603.03380},
  archivePrefix = {arXiv},
  primaryClass = {cs.RO},
  doi = {10.48550/arXiv.2603.03380},
  url = {https://arxiv.org/abs/2603.03380}
}

@inproceedings{libero,
  author = {Liu, Bo and Zhu, Yifeng and Gao, Chongkai and Feng, Yihao and Liu, Qiang and Zhu, Yuke and Stone, Peter},
  title = {{LIBERO}: Benchmarking Knowledge Transfer for Lifelong Robot Learning},
  booktitle = {Advances in Neural Information Processing Systems},
  volume = {36},
  pages = {44776--44791},
  year = {2023},
  doi = {10.52202/075280-1939},
  url = {https://proceedings.neurips.cc/paper_files/paper/2023/hash/8c3c666820ea055a77726d66fc7d447f-Abstract-Datasets_and_Benchmarks.html}
}

@inproceedings{mujoco,
  author = {Todorov, Emanuel and Erez, Tom and Tassa, Yuval},
  title = {{MuJoCo}: A Physics Engine for Model-Based Control},
  booktitle = {2012 IEEE/RSJ International Conference on Intelligent Robots and Systems},
  pages = {5026--5033},
  year = {2012},
  doi = {10.1109/IROS.2012.6386109},
  url = {https://doi.org/10.1109/IROS.2012.6386109}
}

\end{document}